\documentclass[conference]{IEEEtran}
\IEEEoverridecommandlockouts
\usepackage{cite}
\usepackage{amsmath,amssymb,amsfonts}
\usepackage{algorithmic}
\usepackage{algorithm}
\usepackage{graphicx}
\usepackage{textcomp}
\usepackage{subcaption}
\usepackage{verbatim}
\usepackage{xcolor}
\usepackage{balance}
\usepackage{tabularx,booktabs}
\def\BibTeX{{\rm B\kern-.05em{\sc i\kern-.025em b}\kern-.08em
    T\kern-.1667em\lower.7ex\hbox{E}\kern-.125emX}}
\makeatletter
\def\ps@headings{
\let\@oddhead\@empty
\let\@evenhead\@empty
\def\@oddfoot{\@IEEEheaderstyle\hfil\thepage}%
\def\@evenfoot{\@IEEEheaderstyle\thepage\hfil\hbox{}}
}
\def\ps@IEEEtitlepagestyle{
\let\@oddhead\@empty
\let\@evenhead\@empty
\def\@oddfoot{\footnotesize DOI: 10.1109/BigData59044.2023.10386795 ~\copyright 2024 IEEE.  Personal use of this material is permitted. Permission from IEEE must be obtained for all other uses.\hfill}%
\let\@evenfoot\@empty
}
\makeatother
\begin{document}

\title{Self-supervised visual learning for analyzing firearms trafficking activities on the Web\\
\thanks{The research leading to these results has received funding from the European Union's Horizon Europe research and innovation programme under grant agreement No 101073876 (Ceasefire).}}

\author{\IEEEauthorblockN{1\textsuperscript{st} Sotirios Konstantakos}
\IEEEauthorblockA{\textit{Department of Informatics and Telematics} \\
\textit{Harokopio University of Athens}\\
Athens, Greece \\
skwnstantakos@hua.gr}
\and
\IEEEauthorblockN{2\textsuperscript{nd} Despina Ioanna Chalkiadaki}
\IEEEauthorblockA{\textit{Department of Informatics and Telematics} \\
\textit{Harokopio University of Athens}\\
Athens, Greece \\
it2022112@hua.gr}
\and
\hspace{5em}\IEEEauthorblockN{3\textsuperscript{rd} Ioannis Mademlis}
\IEEEauthorblockA{\hspace{5em}\textit{Department of Informatics and Telematics} \\
\hspace{5em}\textit{Harokopio University of Athens}\\
\hspace{5em}Athens, Greece \\
\hspace{5em}imademlis@hua.gr}
\and
\IEEEauthorblockN{4\textsuperscript{th} Adamantia Anna Rebolledo Chrysochoou}
\IEEEauthorblockA{\textit{Department of Informatics and Telematics} \\
\textit{Harokopio University of Athens}\\
Athens, Greece \\
adamantia.reb@hua.gr}
\and
\hspace{18em}\IEEEauthorblockN{5\textsuperscript{th} Georgios Th. Papadopoulos}
\IEEEauthorblockA{\hspace{18em}\textit{Department of Informatics and Telematics} \\
\hspace{18em}\textit{Harokopio University of Athens}\\
\hspace{18em}Athens, Greece \\
\hspace{18em}g.th.papadopoulos@hua.gr}

}


\maketitle


\begin{abstract}
Automated visual firearms classification from RGB images is an important real-world task with applications in public space security, intelligence gathering and law enforcement investigations. When applied to images massively crawled from the World Wide Web (including social media and dark Web sites), it can serve as an important component of systems that attempt to identify criminal firearms trafficking networks, by analyzing Big Data from open-source intelligence. Deep Neural Networks (DNN) are the state-of-the-art methodology for achieving this, with Convolutional Neural Networks (CNN) being typically employed. The common transfer learning approach consists of pretraining on a large-scale, generic annotated dataset for whole-image classification, such as ImageNet-1k, and then finetuning the DNN on a smaller, annotated, task-specific, downstream dataset for visual firearms classification. Neither Visual Transformer (ViT) neural architectures nor Self-Supervised Learning (SSL) approaches have been so far evaluated on this critical task. SSL essentially consists of replacing the traditional supervised pretraining objective with an unsupervised pretext task that does not require ground-truth labels. This paper evaluates a common CNN and a typical ViT architecture in combination with different SSL methods, comparing them against each other and against supervised pretraining in terms of downstream classification accuracy. Additionally, ``CrawledFirearmsRGB" is introduced as a new, challenging image dataset for visual classification of firearms and other concepts related to on-line criminal networks. Finally, a new mixed pretraining objective is formulated that combines SSL and whole-image classification, under a multitask learning setting. The experimental results, indicate the superiority of certain SSL pretraining methods that cooperate well with the ViT architecture, even when the pretraining dataset is of a scale similar or identical to that of CrawledFirearmsRGB, despite the fact that no ground-truth labels are exploited.
\end{abstract}

\begin{IEEEkeywords}
Firearms, Security, Open-Source Intelligence, Dark Web, Deep Neural Networks, Image Recognition, Self-Supervised Learning
\end{IEEEkeywords}

\section{Introduction}
Accurate automatic classification of RGB images depicting firearms is a crucial task for a variety of applications, including public security and intelligence gathering. It can power effective firearms detection systems for use in airports, courthouses, and other public places, assist law enforcement organizations in the investigation of firearms-related crimes, as well as aid intelligence agencies in tracking the movement of firearms and identify potential threats or trafficking networks \cite{Mademlis2024}.

Open-source intelligence gathering regarding illicit firearms may involve crawling and analysis of images from the Internet on a massive scale, essentially rendering efficient visual firearms classification a Big Data task. Unlike public space security applications, where imaging technologies such as X-ray scans may be more appropriate for identifying concealed weapons \cite{Batsis2023} \cite{Mademlis2023batsis}, regular RGB images of firearms are the ones most commonly found with open source intelligence from the Internet. Forums, social media posts, Dark Web marketplaces, and even commercial e-shops can be utilized as image sources, since on-line depictions of firearms may be related to illegal activity.

Image classification in general is a well-known task in computer vision \cite{papadopoulos2016deep}. However, automated visual firearms classification can be challenging due to the large diversity of firearm types, models, and appearances (e.g., due to viewpoint, background or illumination variations). Additionally, firearms images can be occluded or partially obscured, making them difficult to identify. Deep Neural Networks (DNNs) have achieved state-of-the-art performance in similar cases and have indeed been applied to firearms classification. Yet, DNNs typically require large amounts of labeled training data; this can be a challenge for firearms classification, as relevant domain-specific labeled image datasets are often limited in size and expensive to collect.

Self-Supervised Learning (SSL) is a machine learning paradigm that allows DNNs to learn useful image features from unlabeled data. With SSL, DNNs are first pretrained to solve auxiliary tasks designed to teach the model how to extract general-purpose features. These so-called ``pretext tasks" do not rely on the existence of ground-truth annotation/class labels \cite{Jing2020} \cite{Patrona2021}. After pretraining, the DNN is typically finetuned on the desired supervised downstream task, such as firearms classification, using the available annotated data that may be limited in size. Notably, this setting differs from the traditional transfer learning setting, where DNNs are first pretrained in a regular supervised manner on a generic large-scale image dataset (e.g., on ImageNet-1k \cite{deng2009imagenet} for whole-image classification) and then finetuned on the desired downstream task.

Despite the boost in final test-stage accuracy that has been observed in various applications when SSL pretraining is applied before downstream task-specific training, the SSL paradigm has not yet been studied and evaluated in the domain of visual firearms classification. Yet, it is a task that would likely benefit significantly from SSL pretraining, due to the lack of relevant large-scale annotated datasets. Moreover, existing methods limit themselves to the use of Convolutional Neural Networks (CNNs) and newer neural architectures, such as Vision Transformers, have not been exploited.

In an attempt to fill this gap, this paper makes the following contributions to the field of visual firearms classification:
\begin{itemize}
\item It provides a comprehensive comparison of four different, recent SSL pretraining algorithms (SimCLR \cite{chen2020simple}, DeepClusterV2 \cite{caron2018deep}, DINO \cite{caron2021emerging}, and MAE \cite{he2022masked}) against two conventional supervised approaches: pretraining on a) ImageNet-1k \cite{deng2009imagenet}, or b) on the significantly smaller ImageNet-100  dataset \cite{zhao2020maintaining}.
\item It evaluates a new, mixed pretraining scheme that combines a self-supervised objective with a traditional supervised objective, under a multitask learning setting.
\item It employs a new large-scale dataset, i.e., ``CrawledFirearmsRGB", containing 25,000 manually collected and annotated RGB images that depict firearms and related concepts.
\item It evaluates a Vision Transformer neural architecture on this application domain for the first time.
\item It experimentally shows that SSL pretraining non-negligibly improves the accuracy of visual firearms classification and that its downstream performance depends less on the size/scale of the pretraining dataset, compared to traditional supervised alternatives.
\end{itemize}

The remainder of the paper is organized as follows. Section \ref{sec:RelatedWork} overviews the recent literature on visual firearms classification. Section \ref{sec:Methods} briefly presents the specific SSL pretraining approaches that are being empirically assessed in this paper, as well as the proposed mixed pretraining scheme. Section \ref{sec:Evaluation} outlines the experimental evaluation process, presents the employed dataset and discusses the obtained results. Section \ref{sec:Conclusions} concludes the preceding discussion, identifying the implications of these findings and directions for future research.

\section{Related Work}
\label{sec:RelatedWork}
A lot of relevant research has focused on visual firearms classification on RGB images, with DNNs comprising the state-of-the-art. Identical methods can be used for various types of image sources (e.g., World Wide Web, surveillance footage, etc.), including video sources. For instance, the method in \cite{singleton2018gun} leverages the fast MobileNet CNN architecture \cite{howard2017mobilenets} for real-time identification of handguns in surveillance video (e.g., CCTV footage). The approach in \cite{gelana2019firearm} builds upon this line of research, by pruning the amount of video content that needs to be subsequently analyzed by the CNN. This is done via consecutively applying video key-point detection, elimination of overly clustered key-points and background removal, thus resulting in efficiency and accuracy gains.

The method in \cite{augdacs2021deep} evaluates the effect of transfer learning on firearms recognition in RGB images, in a comparison between pretrained AlexNet \cite{krizhevsky2012imagenet} and VGG16/VGG19 CNN architectures \cite{simonyan2014very}. In order to facilitate the experimental process, the authors also create a comprehensive annotated dataset by merging commonly used, public image datasets depicting various weapons and additional data crawled from the Internet. The positive effect of supervised pretraining on generic, large-scale datasets and of training data augmentation is showcased.

Similarly, the Inception V3 \cite{szegedy2016rethinking} CNN architecture is utilized in \cite{cardoza2022convolutional} for handgun classification in images. The study indicates the accuracy gains induced by a more meticulous search for optimal CNN hyperparameter values. The method in \cite{kaya2021detection} employs the VGG-16 and ResNet-50/ResNet-101 \cite{he2016deep} CNN architectures to recognize a more wide range of weapon types, such as assault rifles, bazookas, grenades, hunting rifles, knives, pistols, and revolvers. To achieve this, it introduces a relevant image dataset. Finally, in \cite{pang2022image}, a custom, fast and simple CNN is developed for weapon recognition in images, instead of generic, well-known neural architectures.

Overall, the performance of the above methods heavily depends on the size of domain-specific annotated datasets, as they are trained under a fully supervised setting. Severe overfitting is the typical negative outcome of this situation, due to the large model complexity of DNNs. Although transfer learning by supervised pretraining on large-scale, generic/task-agnostic datasets may reduce overfitting and push the DNN towards learning to extract better features, the domain discrepancy between the pretraining dataset and the task-specific, downstream dataset limits the achievable gains. None of the existing methods for visual firearms classification has exploited SSL in order to overcome this issue.

Besides whole-image classification, firearms or weapon detection has also been attempted on RGB images or videos \cite{brahmaiah2021artificial} \cite{rose2020assessment}. Although object detection falls besides the scope of this paper, it is noteworthy that SSL or Vision Transformers have not been employed there as well.

\section{Overview of Self-Supervised Learning Algorithms}
\label{sec:Methods}
Self-Supervised Learning (SSL) in computer vision consists of two consecutive steps: a) pretraining a neural network for an unsupervised pretext task, and b) finetuning the resulting model on the desired downstream task, in a regular supervised manner. The pretraining dataset may differ from the downstream dataset, i.e., the so-called ``SSL transfer learning setting", or not, i.e., the so-called ``SSL single-dataset setting". When it differs, the pretraining dataset is typically much larger in size than the downstream one. In the single-dataset setting, the ground-truth labels are simply not utilized during the pretraining stage.
In both cases, the DNN that is eventually obtained has learned to extract richer features and better input representations, compared to downstream training from scratch. The traditional alternative to achieve a similar effect is the typical supervised transfer learning setting: a) firstly, pre-training on a large-scale annotated dataset for regular supervised whole-image classification, and b) subsequently, fine-tuning on the downstream dataset.

Four SSL algorithms are selected to be included in this study: SimCLR, DeepClusterV2, DINO and MAE. Two criteria were considered to select them: a) they are all relatively recent and/or state-of-the-art, and b) they are representatives of different families of SSL methodologies. What distinguishes different SSL methods is the exact formulation of their respective pretext task.

\subsection{SimCLR}
SimCLR \cite{chen2020simple} is a method designed for contrastive self-supervised learning. Unlike earlier similar methods \cite{oord2018representation} \cite{he2020momentum}, which often depend on specific architectures or memory banks, SimCLR operates without such requirements.

Contrastive learning is a training approach that brings similar samples closer in the representation space, while pushing dissimilar samples apart. In the context of computer vision, positive pairs are similar images and negative pairs are dissimilar ones. The SimCLR pretext task aims to maximize the similarity of representations for positive pairs and minimize it for negative pairs. Given that no ground-truth semantic labels are utilized, similar images are obtained as differently transformed variants of a single image, while dissimilar ones are essentially distinct images from the training dataset. Such a mechanism compels the DNN to learn to identify and amplify semantically meaningful image features.

It has been shown that data augmentation is instrumental to the performance achieved by SimCLR. Transformations such as cropping, flipping, and color changes are utilized to generate different versions of a single input image. These variants are used as positive pairs during training, facilitating the contrastive learning mechanism.

A key feature of SimCLR is its incorporation of a modifiable nonlinear transformation, implemented as a trainable, fully-connected neural layer. Its primary role is to map these features into an optimized space conducive for contrastive learning, ensuring effective comparison of similarities and differences between image representations. Thus, during training, it finetunes the input representations before their utilization in the contrastive loss computation.

The loss function used by SimCLR for pretraining is the Normalized Temperature-scaled Cross Entropy (NT-Xent):
\begin{equation}
\mathcal{L}_{NT-Xent}(\mathcal{B}) = -\log \frac{\exp(\text{sim}(\mathbf{b}_i, \mathbf{b}_j) / \tau)}{\sum_{k=1}^{2B} \mathbb{1}_{[k \neq i]} \exp(\text{sim}(\mathbf{b}_i, \mathbf{b}_k) / \tau)},
\end{equation}
\noindent where $\text{sim}(\cdot)$ is the cosine similarity, $\tau$ is a scalar temperature hyperparameter and $\mathcal{B}$ is the set of $B$ images in the current training mini-batch. $\mathbf{b}_i$ and $\mathbf{b}_j$ are the latent representations of two differently augmented transformations of the in-batch image currently utilized as an anchor. Each image in the mini-batch is similarly augmented twice, resulting in $2B$ representations for negative sampling. The indicator function $\mathbb{1}_{[k \neq i]}$ is 1 when $k \neq i$ and 0 otherwise.

\begin{algorithm}
\label{alg::SimCLRpseudo}
\caption{Pretraining algorithm of SimCLR.}
\begin{algorithmic}[1]
\STATE Initialize DNN parameters.
\WHILE{not converged}
    \STATE Take an image and apply two distinct transformations to get two views.
    \STATE Use DNN to extract features for each view.
    \STATE Apply learnable nonlinear transformation to the features.
    \STATE Compute $\mathcal{L}_{NT-Xent}$.
    \STATE Backpropagate the error from $\mathcal{L}_{NT-Xent}$ to update DNN parameters.
\ENDWHILE
\end{algorithmic}
\end{algorithm}

SimCLR shows improved performance with increased mini-batch sizes and extended training iterations, when compared to standard supervised learning models. Its pseudocode is shown in Algorithm 1.

\subsection{DINO}
DINO \cite{caron2021emerging} (Distillation-based Inference NOt needing labels) is an SSL method designed for Vision Transformers (ViTs), although it can also be applied to CNNs. DINO applied to ViTs demonstrates a unique capability: the features extracted by the pretrained DNN model often encompass explicit semantic segmentation information. This characteristic is not as evident in CNNs or in ViTs pretrained regularly for whole-image classification, suggesting that DINO has a distinctive method of encoding semantic information and that it is particularly aligned with the architecture and operation of ViTs. The integration of momentum encoders \cite{he2020momentum} and multi-crop training \cite{caron2020unsupervised} has been observed to improve the training process in terms of both speed and accuracy.

With DINO pretraining, the DNN learns to iteratively refine its predictions, drawing from its own previously gained knowledge. Thus, a central aspect of DINO is its self-distillation mechanism: a student network is trained to match the soft outputs of a more stable teacher network \cite{Hinton2015}. The two networks are architecturally identical and process differently augmented/transformed views of an input image. The pretext task objective is for the two DNNs to generate similar outputs for these different views, so that it does not simply memorize data but understands underlying features. Thus, the training process attempts to make the student's predictions consistent with the soft outputs of the teacher.

The two networks are parameterized differently, with the teacher parameters being iteratively updated as an exponential moving average of the student parameters, ensuring a slow, consistent update. The student parameters are updated by a conventional combination of error back-propagation and a variant of Stochastic Gradient Descent. Let us assume that $\mathbf{B}^g_1, \mathbf{B}^g_2$ are two differently augmented ``global" crops of a single training image, i.e., covering a significant portion of the original image's area, while $k$ additional, differently augmented ``local" crops, i.e., covering a substantially smaller portion of the original image's area, are independently and randomly derived. $\mathcal{V}$ is the set of all $k+2$ views obtained from a single training image, which includes $\mathbf{B}^g_1$ and $\mathbf{B}^g_2$ among its elements. All $k+2$ views are passed through the student network $S(\cdot)$, but only the two global views are passed through the teacher network $T(\cdot)$. Then the loss function employed for DINO pretraining is the following one:
\begin{equation}
    \mathcal{L}_{DINO} = \sum_{\substack{\mathbf{B}_i\in\{\mathbf{B}^g_1, \mathbf{B}^g_2\}}} \sum_{\substack{\mathbf{B}_j\in\mathcal{V} \\ \mathbf{B}_j\neq\mathbf{B}_i}} \mathcal{L}_{CE}(T(\mathbf{B}_i),S(\mathbf{B}_j)),
\end{equation}
\noindent where $\mathcal{L}_{CE}$ is the Cross-Entropy loss, while the outputs of $S(\cdot)$ and $T(\cdot)$ are softmax distributions. The objective of $\mathcal{L}_{DINO}$ is to align the softmax-derived output probability distributions of the student network with those of the teacher network for the two data views.

\begin{algorithm}
\caption{Pretraining algorithm of DINO.}
\label{alg::DINOpseudo}
\begin{algorithmic}[1]
\STATE Initialize student (ViT) and teacher parameters.
\WHILE{not converged}
    \STATE Take an image and apply multi-crop augmentation to get multiple views.
    \STATE Use student ViT to extract features for each view.
    \STATE Compute $\mathcal{L}_{\text{DINO}}$.
    \STATE Back-propagate the error from the $\mathcal{L}_{DINO}$ loss to update student ViT parameters.
    \STATE Update teacher parameters using the momentum encoder with student parameters.
\ENDWHILE
\end{algorithmic}
\end{algorithm}

After pretraining has been completed, the teacher network can be discarded. The DINO pseudocode is shown in Algorithm 2.

\subsection{MAE}
MAE \cite{he2022masked} (Masked AutoEncoders) is an SSL method for computer vision where the pretext task consists in forcing the DNN to reconstruct parts of its input image that have been masked. This stands in contrast to traditional autoencoders, which try to reproduce their entire input. The method assumes an asymmetric Encoder-Decoder neural architecture and extensive input image masking, so that semantically meaningful representations can be learned without any ground-truth labels. Masking up to 75\% of the input pixels makes the task challenging and, thus, the DNN is compelled to learn rich image representations.

The Encoder processes the visible image regions and ignores the masked ones. The Decoder, on the other hand, attempts to reconstruct the masked regions using the information relayed by the Encoder. By this design, the Decoder predicts the masked portions based solely on the visible content. This clear separation of roles in the architecture facilitates effective training and accurate reconstruction of the masked regions.

The MAE pretraining objective directly penalizes discrepancies between the reconstructed portions of the image (from the DNN's output) and the corresponding regions in the original image \( \mathbf{X} \). This ensures that the DNN accurately predicts the pixel values of the masked patches, thus enhancing the quality of its reconstructions. The employed loss function is the MSE between the DNN's output \(\mathbf{P} \in \mathbb{R}^{W \times H \times 3}\) and the original image \(\mathbf{Y} \in \mathbb{R}^{W \times H \times 3}\): 

\begin{equation}
\mathcal{L}_{MAE} = \mathcal{L}_{MSE}(\mathbf{P}, \mathbf{Y}),
\end{equation}
\noindent where
\begin{equation}
\mathcal{L}_{MSE}(\mathbf{v}, \hat{\mathbf{v}}) = \frac{1}{L} \sum_{i=1}^{L} (v_i - \hat{v}_i)^2,
\end{equation}

\noindent assuming that $\mathbf{v} \in \mathbb{R}^L$ and $\hat{\mathbf{v}} \in \mathbb{R}^L$ are the pseudo-ground-truth and the predicted/reconstructed vector.

However, in contrast to typical applications of $\mathcal{L}_{MSE}$, only the pixels corresponding to masked image regions are evaluated within the loss function; the unmasked/visible input patches do not contribute to loss calculations. After pretraining has been completed, the Decoder can be discarded. It has been experimentally shown that MAE can outperform regular supervised pretraining in transfer learning situations. However, it is far from certain that it is universally useful to all application domains, such as visual firearms classification.

\begin{algorithm}
\caption{Pretraining algorithm of MAE.}
\label{alg::MAEpseudo}
\begin{algorithmic}[1]
\STATE Initialize Encoder and Decoder parameters.
\WHILE{not converged}
    \STATE Randomly mask regions of the input image to get a severely masked image.
    \STATE Use Encoder on visible regions of the masked image to get its latent representation.
    \STATE Use Decoder on the latent representation and mask tokens to reconstruct the original image.
    \STATE Compute $\mathcal{L}_{MAE}$.
    \STATE Back-propagate the error from the $\mathcal{L}_{MAE}$ loss to update Encoder and Decoder parameters.
\ENDWHILE
\end{algorithmic}
\end{algorithm}

The MAE pseudocode is shown in Algorithm 3.

\subsection{DeepClusterV2}
DeepCluster \cite{caron2018deep} is an SSL method where the pretext tasks consists of training for supervised classification, using pseudolabels from a previous unsupervised clustering stage. These two stages alternate until model convergence.

Initially, the DNN infers features from the training data, which are fed into the K-Means algorithm \cite{Lloyd1982}, in order to generate distinct cluster assignments. The number of clusters $K$ is a fixed hyperparameter. These assignments, or cluster IDs, are then exploited as pseudolabels that drive the subsequent supervised training stage. During the latter, the DNN's objective shifts to predicting these dynamically generated pseudolabels, leading to parameter adjustments that refine the features to be used as input for the upcoming clustering iteration. In short, the evolution of the clustering feature space is intermingled with the DNN training progress.

The loss function utilized for the classification training stage is the regular Cross-Entropy loss:

\begin{equation}
\mathcal{L}_{\text{DC}}(\mathbf{y}^i, \mathbf{p}^i) = -\sum_{c=1}^{C} y^i_c \log\left(softmax(\mathbf{p}^i_c)\right),
\end{equation}


\noindent where $\mathbf{y}^i$ represents the pseudolabel (cluster assignment vector by K-Means) for the \(i\)-th training sample \(\mathbf{x}_i\), taken from a total of \(N\) samples, while $\mathbf{p}^i$ is the respective predicted vector at the final output of the DNN. The detailed procedure of DeepCluster is shown in Algorithm 4. The method has been improved with DeepClusterV2.

\begin{algorithm}
\label{alg::DCpseudo}
\caption{Pretraining algorithm of DeepCluster.}
\begin{algorithmic}[1]
\STATE Initialize DNN parameters.
\WHILE{not converged}
    \STATE Use DNN to extract features from the dataset.
    \STATE Apply K-Means clustering on the extracted features to get cluster assignments.
    \STATE Use cluster assignments as pseudolabels and train DNN with loss $\mathcal{L}_{DC}$.
    \STATE Back-propagate the error from $\mathcal{L}_{DC}$ to update DNN parameters.
\ENDWHILE
\end{algorithmic}
\end{algorithm}

\subsection{Mixed Pretraining}
Besides employing the 4 above-described SSL methods and comparing them with regular supervised pretraining, using visual firearms classification as the downstream task, a new mixed pretraining scheme is also introduced in this paper. It consists of combining an SSL pretext task with regular supervised classification during the pretraining stage, in order for the DNN to leverage the advantages of both and to learn richer features. Evidently, it can only be used when the pretraining dataset is annotated with ground-truth labels.

This scheme can be implemented by appending two parallel heads to the DNN, where the first/second one generates the desired output of the SSL/supervised task, respectively. Pretraining proceeds using a composite objective function, in a multitask learning approach. After convergence, the two parallel heads are simply replaced with a single new one, which is suitable for the downstream task. Then, supervised finetuning for the latter one proceeds using the pretrained DNN as a pre-initialized backbone.

The composite objective function is the following one:

\begin{equation}
\mathcal{L}_{\text{mixed}} = \omega_1 \mathcal{L}_{supervised} + \omega_2 \mathcal{L}_{SSL},
\label{eq::mixed}
\end{equation}

\noindent where $\mathcal{L}_{supervised}$ and $\mathcal{L}_{SSL}$ are the loss terms for the supervised task (e.g., a cross-entropy loss for classification) and for the employed SSL task, respectively. Obviously, each individual loss term is computed using the output of the respective neural head. The scalar weights $\omega_1$ and $\omega_2$ are manually set hyperparameters that determine the desired balance between the two individual objectives.



\section{Experimental Evaluation}
\label{sec:Evaluation}
This section presents the experimental setup utilized for assessing the efficacy of the methods presented in Section \ref{sec:Methods}, along with the evaluation results.
\subsection{Datasets}
One public RGB image dataset depicting firearms has been introduced so far \cite{perez2020object}. Despite its merits, it is limited in scale (number of images, number of classes) and does not include additional semantic classes that are relevant to firearms-related illegal activities, but are not weapons themselves (e.g., drugs). On the other hand, it does contain irrelevant classes of handheld objects that do not in fact relate to criminal behaviour. Due to these limitations, the introduced, so-called ``CrawledFirearmsRGB" dataset has been manually assembled and annotated. In the conducted experiments, it mainly serves as the downstream finetuning dataset for the transfer learning setup. However, in the single-dataset setup it is also exploited for pretraining.

\subsubsection{The CrawledFirearmsRGB Dataset}
After preliminary, small-scale, manual image collection and annotation/labelling, CrawledFirearmsRGB was specified as having 23 distinct classes. Based on this, an automated, generic Web crawler was developed to ensure systematic and efficient data retrieval. Subsequently, a distinct set of images were crawled specifically from the Reddit on-line platform, in order to enhance the dataset. Post-acquisition processes, such as categorization, cross-referencing for potential similarities, and data annotation, were executed manually in all cases to uphold data veracity.

\begin{figure*}[ht]
    \centering
    \begin{subfigure}[b]{0.22\textwidth}
        \includegraphics[width=\textwidth]{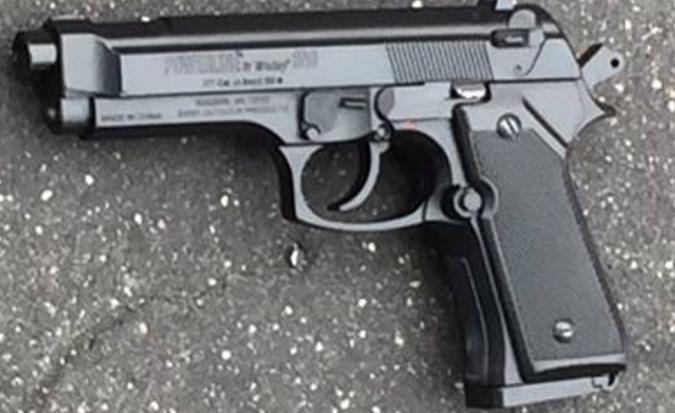}
    \end{subfigure}
    \hfill
    \begin{subfigure}[b]{0.22\textwidth}
        \includegraphics[width=\textwidth]{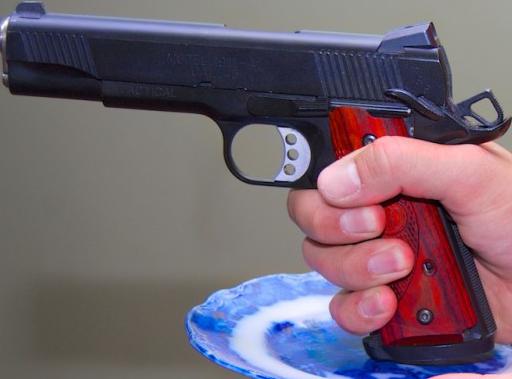}
    \end{subfigure}
    \hfill
    \begin{subfigure}[b]{0.22\textwidth}
        \includegraphics[width=\textwidth]{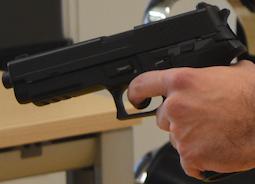}
    \end{subfigure}
    \hfill
    \begin{subfigure}[b]{0.15\textwidth}
        \includegraphics[width=\textwidth]{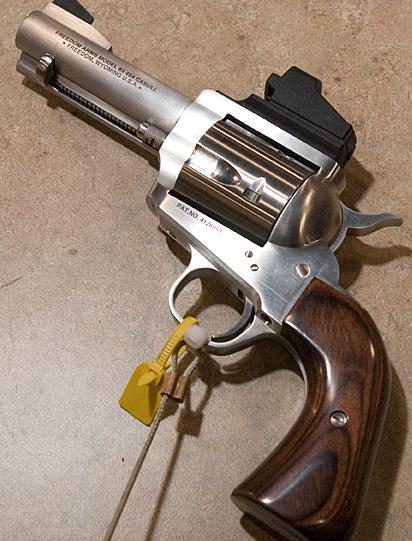}
    \end{subfigure}
    \caption{Sample images from the dataset in \cite{perez2020object}.}
    \label{fig:dataset1}
\end{figure*}

The dataset in \cite{perez2020object} focuses primarily on knives, handguns and similar handled objects; relevant images are depicted in Fig. \ref{fig:dataset1}. In contrast, CrawledFirearmsRGB encompasses a more extensive range of classes that not only differentiate between distinct types of firearms, but also cover various other types of crime-related items besides weapons (e.g., drugs, full-face hoods, etc.). To this end, the assembled dataset is closer to real-world analysis/application cases, i.e., involving diverse visual content (and not only manually selected types of firearms). Hence, the dataset is meant to facilitate automated analysis of World Wide Web content that may hint at any criminal activities linked to the use of firearms. Consistency in the data collection and manual annotation processes, across the identified classes, ensures a high degree of accuracy and reliability in the data.

\begin{table*}
\centering
\begin{tabularx}{\linewidth}{|c|l|c|X|}
\hline
\textbf{Class ID} & \textbf{Name} & \textbf{\# of images} & \textbf{Description} \\
\hline
1 & Bomb & 1245 & Explosive devices that can cause significant destruction. \\
2 & Rifle & 2005 & Long-barreled firearms designed for accurate shooting. \\
3 & Revolver & 1005 & Handguns with a rotating cylinder holding ammunition. \\
4 & Rocket & 1073 & Projectiles driven by engines, often used as weapons. \\
5 & Shotgun & 4381 & Designed for close-range shooting with spread ammunition. \\
6 & Knives & 1842 & Sharp tools often used for cutting or as weapons. \\
7 & PCP airguns & 1067 & Utilize pre-compressed air to propel pellets or BBs. \\
8 & Pills (drugs) & 1061 & Solid dosage forms of medication, sometimes used illicitly. \\
9 & Pistols & 3349 & Short-barreled firearms designed for one-handed use. \\
10 & Weeds & 683 & Plants with potential narcotic properties. \\
11 & Seeds (drugs) & 837 & Related to drug-producing plants. \\
12 & Bullet box & 2081 & Containers used for storing firearm ammunition. \\
13 & Bullets & 1017 & Projectile components fired from firearms. \\
14 & Bow and arrow & 69 & Ancient projectile weapons for hunting and combat. \\
15 & Injectable drugs & 155 & Liquid drugs introduced into the body via syringes. \\
16 & Powder (drugs) & 525 & Ground or pulverized drug substances. \\
17 & Military clothing & 237 & Designed for combat scenarios with protection and camouflage. \\
18 & Full-face hoods & 761 & Worn to conceal one’s identity in tactical situations. \\
19 & Accessories & 207 & Pertaining to weapon-related supplementary items. \\
20 & Blades & 115 & Thin-edged tools or weapons, often sharper than knives. \\
21 & Gun cases & 470 & Protective storage solutions for firearms. \\
22 & Gun storage & 460 & Dedicated spaces or containers for safekeeping weapons. \\
23 & Weapon magazines & 355 & Components that store and feed ammunition into firearms. \\
\hline
\end{tabularx}
\caption{Contents of the CrawledFirearmsRGB dataset.}
\label{tab:Dataset}
\end{table*}

\begin{table}
\centering
\caption{Experimental results for the ViT architecture in the transfer learning setup. CrawledFirearmsRGB is the downstream finetuning dataset in all cases.}
\begin{tabular}{@{}p{2cm} p{2.5cm} c@{}}
\toprule
\textbf{Experiment} & \textbf{Pretraining Dataset} & \textbf{Classification Accuracy (\%)} \\
\midrule
Supervised  & ImageNet-100  & 66.81 \\
Supervised  & ImageNet-1k  & 68.32 \\
MAE         & ImageNet-100  & 67.89 \\
SimCLR      & ImageNet-100  & 69.18 \\
DeepClusterV2 & ImageNet-100  & 67.40 \\
\textbf{DINO} & ImageNet-100  & \textbf{71.94} \\
Mixed (DINO+Supervised) & ImageNet-100 & 71.71 \\
\bottomrule
\end{tabular}
\label{tab::ViTTransfer}
\end{table}

\begin{table}
\centering
\caption{Experimental results for the ResNet-50 architecture in the transfer learning setup. CrawledFirearmsRGB is the downstream finetuning dataset in all cases.}
\begin{tabular}{@{}p{2cm} p{2.5cm} c@{}}
\toprule
\textbf{Experiment} & \textbf{Pretraining Dataset} & \textbf{Classification Accuracy (\%)} \\
\midrule
Supervised  & ImageNet-100  & 68.53 \\
Supervised  & ImageNet-1k   & 69.67 \\
MAE         & ImageNet-100  & 65.75 \\
\textbf{SimCLR}      & ImageNet-100  & \textbf{71.21} \\
DeepClusterV2 & ImageNet-100  & 69.47 \\
DINO        & ImageNet-100  & 69.86 \\
\bottomrule
\end{tabular}
\label{tab::ResNetTransfer}
\end{table}

\begin{table}
\centering
\caption{Experimental results for the ViT architecture in the single-dataset setup.}
\begin{tabular}{@{}p{3.5cm} c@{}}
\toprule
\textbf{Self-Supervised Pretraining (Method)} & \textbf{Classification Accuracy (\%)} \\
\midrule
No pretraining (random initialization)  & 69.78 \\
SimCLR      & 69.23 \\
DeepClusterV2 & 66.29 \\
\textbf{DINO}  & \textbf{72.31} \\
MAE &  66.82 \\
\bottomrule
\end{tabular}
\label{tab::ViTSingle}
\end{table}

\begin{table}
\centering
\caption{Experimental results for the ResNet-50 architecture in the single-dataset setup.}
\begin{tabular}{@{}p{3.5cm} c@{}}
\toprule
\textbf{Self-Supervised Pretraining (Method)} & \textbf{Classification Accuracy (\%)} \\
\midrule
No pretraining (random initialization)  & 68.24 \\
\textbf{SimCLR}      & \textbf{70.53} \\
DeepClusterV2 & 68.57 \\
DINO & 69.64 \\
MAE &  65.31 \\
\bottomrule
\end{tabular}
\label{tab::ResNetSingle}
\end{table}

Table \ref{tab:Dataset} summarizes the characteristics of the CrawledFirearmsRGB dataset. As it can be seen, it has a total of 25000 images with a high degree of imbalance in the size of the various classes, a property which reflects the real-world frequency of appearance of relevant categories. Moreover, there is significant intraclass appearance variance, in terms of object design, angle, lighting, or even context. The dataset was randomly split, according to a 70/15/15 ratio, into a training/validation/test set of 17,500/3,750/3,750 images, respectively. Overall, CrawledFirearmsRGB is a challenging and non-trivial dataset for whole-image classification, which is suitable for training data-hungry models, such as DNNs.

\subsubsection{Transfer Learning Setup}
Under the transfer learning setup, the large-scale, annotated ImageNet-100 dataset \cite{zhao2020maintaining} is employed for pretraining and, subsequently, CrawledFirearmsRGB is utilized for downstream finetuning. This is repeated separately for all 6 competing pretraining approaches: SimCLR, DINO, MAE, DeepClusterV2, mixed and supervised (regular whole-image classification). Among the 4 SSL approaches, only the best-performing one in the downstream task was utilized for the mixed pretraining setup. Obviously, the ground-truth annotations were not utilized when pretraining with the 4 pure SSL approaches.

Besides the above variants, the pretraining scenario most commonly encountered in practical situations, i.e., supervised pretraining on ImageNet-1k \cite{deng2009imagenet}, which is one order of magnitude larger in size than ImageNet-100, was also separately evaluated in combination with downstream finetuning on CrawledFirearmsRGB. In all datasets, utilized either for pretraining or for finetuning, common data augmentation schemes were uniformly applied, such as random cropping, flipping, and normalization.

\subsubsection{Single-Dataset Setup}
Under the single-dataset setup, one common dataset, i.e., CrawledFirearmsRGB, is used both for pretraining and for finetuning, in two consecutive stages. Obviously, the ground-truth annotations are not exploited during the SSL pretraining stage. Although the traditional supervised pretraining approaches are not directly applicable in this scenario, the downstream accuracy of SSL-pretrained models can be compared against the respective supervised-pretrained approaches of the transfer learning setup.

\subsection{Experimental Results}
The DNN architectures employed in the experimental assessment are both CNN (ResNet-50) and Transformer (ViT) models. The first one is commonly utilized in practical tasks and not particularly demanding resource-wise. ViT has been selected due to the existence of SSL methods that are especially attuned to Transformer architectures (DINO, MAE). Traditional supervised pretraining has been evaluated for both architectures. The code was developed in Python, using the PyTorch framework, and the experiments were conducted on nVidia A100 GPUs. Hyperparameters were optimized using grid search. Regarding the scalar coefficients of the mixed method in Eq. (\ref{eq::mixed}), optimal values were found to be $\omega_1 = 0.45 $ and $\omega_2 = 0.55$.

Tables \ref{tab::ViTTransfer} and \ref{tab::ResNetTransfer} summarize the experimental results for the transfer learning setup. As it can be seen, DINO pretraining on ViT leads to the best downstream accuracy. Thus, DINO was selected as the SSL component of the mixed pretraining scheme, which comes second in performance (after pure DINO). It is notable that both DINO and mixed pretraining on ImageNet-100 surpass in downstream accuracy regular supervised pretraining on the much larger ImageNet-1k. This observation confirms the insight that proper SSL pretext tasks can lead to significantly richer feature extraction without the need for ground-truth labels, thus mitigating the data inefficiency of typical DNN training. Without SSL, there is a notable downstream performance gap between pretraining on ImageNet-1k and on ImageNet-100, in favour of the former, since a larger dataset introduces the model to a more diverse set of image features, instances and nuances. This observation also explains why the mixed pretraining scheme (DINO + supervised under a multitask learning setting) is unable to surpass pure DINO with ImageNet-100. In this sense, the performance of the mixed scheme confirms the much greater degree of dependence of supervised pretraining on the dataset size/scale, in comparison to SSL alternatives. Obviously, these results are specific to the visual firearms recognition task, particularly given the dataset size and diversity of CrawledFirearmsRGB, as it was the only downstream dataset used.

\begin{figure*}[htbp]
    \centering
    \includegraphics[width=\linewidth]{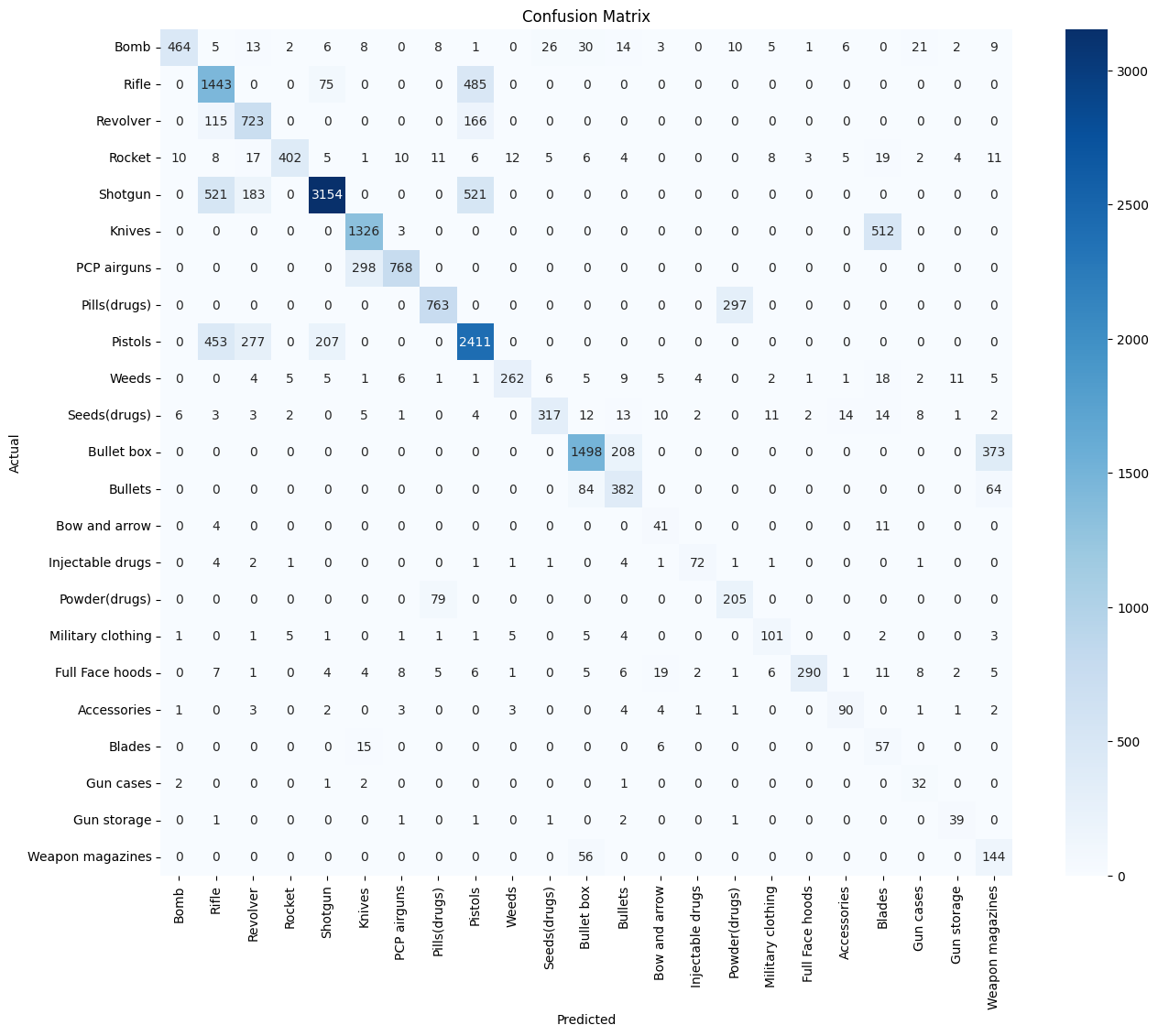}
    \caption{Confusion matrix of DINO on ViT (single-dataset setup).}
    \label{fig:confusion_matrix}
\end{figure*}

Regarding the performance of the various SSL methods, SimCLR also performs well and better than all purely supervised pretraining approaches, even with a ResNet-50 backbone. Most likely, this high degree of effectiveness of SSL methods stems from their ability to unearth patterns/features that might be subdued or overlooked in strictly supervised pretraining, due to the fact that the DNN's training is not confined by potentially limiting interclass boundaries. Instead, it focuses on raw, feature-driven similarities and distinctions between images. During finetuning on CrawledFirearmsRGB, these capabilities translate to a greater discriminatory ability linked to firearm-related nuances.
In contrast, DeepClusterV2 or MAE pretraining on ImageNet-100 are unable to increase downstream classification accuracy compared to supervised pretraining on ImageNet-1k, although they both do lead to small improvements over the respective supervised pretraining on ImageNet-100. Therefore, in an apples-to-apples comparison on ImageNet-100, both DeepClusterV2 and MAE prove effective, but not enough to outperform the traditional ImageNet-1k transfer learning approach. This performance disparity in comparison to SimCLR or DINO suggests that CrawledFirearmsRGB contains certain latent structures or patterns that are particularly amenable to contrastive and self-distillation SSL methods. Furthermore, DINO inherently emphasizes enhancing representational consistency between different transformations of a single image. Given the nature of CrawledFirearmsRGB, where firearms can manifest subtle design variations and appearances based on different conditions or viewing angles, this property becomes pivotal. Essentially, DINO ensures that multiple depictions of similar objects are consistently interpreted, addressing the issue of high intraclass appearance variance found in the dataset.

Tables \ref{tab::ViTSingle} and \ref{tab::ResNetSingle} summarize the experimental results for the single-dataset setup. In this case, only SSL pretraining is meaningful. The results essentially confirm the insights extracted from the transfer learning setup, with DINO (ViT)/SimCLR (ResNet-50) being ranked as the best/second best approach, respectively. This is most likely due to the focus of said pretext tasks on discerning minute differences between image pairs, with firearms images indeed containing subtle distinctions. The comparative underperformance of MAE and DeepClusterV2 suggests that while they can capture the broad strokes of the dataset, they face difficulties in learning the finer distinctions between classes. As a result, MAE performs worse even than training from scratch with random initialization.

It is noteworthy that both SimCLR and DINO pretraining in the single-dataset setup lead to higher downstream accuracy compared to the traditional supervised pretraining approach, with transfer learning from ImageNet-1k. This is despite the fact that the latter is almost two orders of magnitude larger than CrawledFirearmsRGB and that no ground-truth class labels are exploited by SimCLR and DINO. Additionally, it must be underlined that only by using an SSL pretraining method designed for Transformer architectures that the potential of ViT is fully taken advantage of in the transfer learning setup; otherwise, ResNet-50 outperforms ViT in visual firearms classification. The situation is different in the single-dataset setup, where ViT has an advantage over ResNet-50 in training from scratch. Finally, it is interesting that training ViT from scratch in the single-dataset setup using CrawledFirearmsRGB outperforms almost all transfer learning approaches, except DINO and SimCLR on ResNet-50. More generally, the single-dataset setup leads to the best accuracy in visual classification accuracy, since DINO has a slight $+0.37\%$ advantage compared to the transfer learning setup.

The confusion matrix shown in Fig. \ref{fig:confusion_matrix} provides valuable insights into the behaviour of the DINO-pretrained ViT model in the single-dataset setup, which is the overall best one. Noteworthy observations include frequent misclassifications among the classes ``Rifle", ``Shotgun", and ``Pistols". This suggests potential challenges in disentangling the representations of these firearms, which are indeed more similar to each other compared to other classes of CrawledFirearmsRGB. Similarly, there is a trend of misclassification between ``Knives" and ``PCP airguns", as well as between ``Bullets" and ``Bullet box", indicating the presence of closely aligned feature spaces for these classes. A particularly interesting observation is the confusion of ``Bow and arrow" with both ``Blades" and ``Rifle". This could either hint towards the DNN's encoding of shared latent features among these classes, or reflect the challenges posed by the limited number of training samples for certain classes. The interchange between the categories ``Pills (drugs)" and ``Powder (drugs)" suggests a need for further refinement in the capability to differentiate subtle variations in drug forms.

\section{Conclusions}
\label{sec:Conclusions}
This paper presented a comprehensive comparative experimental analysis of various pretraining approaches for the downstream task of visual firearms classification, using the new, challenging ``CrawledFirearmsRGB" dataset that contains 23 classes of firearms and related concepts. Four Self-Supervised Learning (SSL) methods, namely SimCLR, DeepClusterV2, DINO, and MAE, were compared against each other and against the traditional supervised pretraining approaches, using ImageNet-100 and the much larger ImageNet-1k datasets, while both transfer learning and single-dataset experiments were conducted. The results indicate that the pretraining dataset size is crucial for achieving good accuracy in the traditional setup, but SSL methods are able to bypass the need for huge datasets and learn useful representations from significantly smaller ones, without relying at all on ground-truth annotations. Potentially due to the peculiarities of images depicting firearms, with occasionally low interclass and high intraclass variance, DINO and SimCLR rank as the best pretraining approaches when CrawledFirearmsRGB is utilized as the downstream dataset. The newly introduced mixed pretraining scheme, combining DINO with supervised pretraining under a multitask learning setting, outperforms all competing approaches in the transfer learning setup for visual firearms classification, except pure DINO. This implies that pretraining with supervised whole-image classification on datasets at the scale of ImageNet-100 leads to no additional representation learning benefits compared to DINO. One likely reason is the need of traditional supervised pretraining for very large-scale datasets, but the specific properties of the visual firearms classification task and of the CrawledFirearmsRGB dataset may also contribute to this outcome.

Future research directions for extending this work are varied. First, more extensive evaluation with different firearms image datasets of various sizes could be conducted. Additionally, the mixed pretraining scheme can be evaluated using datasets at the scale of ImageNet-1k, in order to identify whether the supervised learning component starts contributing positively and synergistically to DINO after a certain dataset size threshold.

\section*{Acknowledgment}
The research leading to these results has received funding from the European Union's Horizon Europe research and innovation programme under grant agreement No 101073876 (Ceasefire). This publication reflects only the authors views. The European Union is not liable for any use that may be made of the information contained therein.

The authors would like to thank Anastasios Alexiadis and Maria Makrynioti from the Centre for Research \& Technology, Hellas (CERTH), who aided in the collection of Reddit images.

\balance
\bibliography{Bib.bib}
\bibliographystyle{IEEEtran}

\vspace{12pt}
\color{red}

\end{document}